\documentclass{article}

\usepackage[utf8]{inputenc}
\usepackage[T1]{fontenc}
\usepackage[a4paper,margin=2.5cm]{geometry}
\usepackage{amsmath}
\usepackage{amssymb}
\usepackage{bm}
\usepackage{graphicx}
\usepackage{booktabs}
\usepackage{algorithm}
\usepackage{algorithmic}    
\usepackage[round,authoryear]{natbib}
\usepackage{url}
\usepackage{authblk}
\usepackage[colorlinks=true,allcolors=blue]{hyperref}

\title{KPPax: preprocess differentiable kinetic ODE for machine learning in atmospheric chemistry}

\author[1,3]{Wenqing Peng\thanks{Corresponding author: \href{mailto:wenqing.peng@helsinki.fi}{wenqing.peng@helsinki.fi}}}
\author[1,2,3]{Michael Boy}
\author[2,3]{Zhi-Song Liu}

\affil[1]{Institute for Atmospheric and Earth Systems Research, University of Helsinki, Helsinki, Finland}
\affil[2]{LUT University, Department of Computational Engineering, Lahti, Finland}
\affil[3]{Atmospheric Modelling Centre Lahti (AMC-Lahti), Lahti, Finland}

\date{}

\begin{document}

\maketitle

\begin{abstract}
Atmospheric chemical mechanisms are commonly distributed in the language of the Kinetic PreProcessor (KPP), which generates efficient Fortran or C code but does not fit into the differentiable programming frameworks in which machine-learning models are developed. We present KPPax, a kinetic preprocessor that generates differentiable, just-in-time compiled, and vectorisable chemistry code in JAX from mechanisms written in the KPP language. KPPax evaluates the rate law with sparse operations in log space, which keeps it differentiable, and integrates the stiff system with the implicit solvers and the forward- and reverse-mode adjoints of the \texttt{diffrax} library, so that exact sensitivities are obtained without hand-derived adjoint code. We demonstrate two applications with the POLLU benchmark mechanism. In the first, the Fisher information matrix computed from an automatically differentiated Jacobian predicts, before any fit, which of the 25 rate coefficients a flow-tube-like design can identify, and noisy synthetic estimation confirms the predicted uncertainties of the nine identifiable rate coefficients to within about 30\% while revealing clusters of fast reactions of which only ratios are constrained. In the second, a neural-network correction of an emission inventory is trained offline on per-scenario 4D-Var retrievals and then fine-tuned online through the chemical model, which reduces the error of the biogenic correction by a factor of 2.4--5 and recovers information on low-temperature emissions that the individual retrievals lose. KPPax makes KPP mechanisms directly usable for parameter estimation and hybrid physics--machine-learning modelling.
\end{abstract}

\section{Introduction}
\label{sec:intro}

Chemicals and aerosols in the ambient air affect human health, weather, and climate \citep{seinfeld2016atmospheric}. Their atmospheric concentrations vary strongly in space and time, governed by emission, chemical transformation, aerosol formation, transport by meteorological conditions, and eventual removal by deposition \citep{Brasseur_Jacob_2017}. Comprehensive atmospheric chemical mechanisms, such as the near-explicit Master Chemical Mechanism \citep{jenkin1997tropospheric, saunders2003protocol}, contain thousands of species coupled through more than ten thousand reactions, and hand-coding the resulting stiff ODE system and its Jacobian is tedious and error-prone.

The Kinetic PreProcessor (KPP) automates this process \citep{damian2002kinetic}. Given a chemical mechanism specified in a dedicated description syntax, the KPP language, KPP parses the reaction list with a Flex/Yacc scanner--parser and generates Fortran or C code that evaluates the time derivative of the concentrations and its Jacobian. The generated code exploits the sparsity of the Jacobian through species reordering and sparse linear algebra routines, which keeps the implicit time stepping required by stiff chemistry efficient, and it is paired with a suite of stiff numerical integrators, including Rosenbrock, BDF, and Runge--Kutta methods. KPP also generates direct-decoupled and adjoint sensitivity code, which efficiently computes derivatives of the model solution with respect to rate coefficients, initial concentrations, and other parameters. These sensitivities support applications such as parameter estimation, uncertainty analysis, and variational data assimilation \citep{sandu2003direct}. This tight coupling between mechanism parsing, sparse Jacobian generation, and forward and adjoint sensitivity code generation has made KPP a standard tool in atmospheric chemistry modelling, but it also means that the resulting code inherits the constraints of the Fortran and C environments and of the CPU hardware for which it was designed.

Other tools to preprocess and solve chemical kinetics have been developed for other software and hardware environments. \texttt{pykpp} \citep{henderson_pykpp} is a preprocessor written in Python, which parses mechanisms written in the KPP language and solves the kinetic ODE with SciPy solvers. PyBox generates 0-dimensional box models that include both atmospheric chemistry and aerosol processes, in which the chemistry is solved by the Assimulo ODE solver \citep{topping2018pybox}, and PyCHAM reuses the mechanism parsing of PyBox but solves the ODE with SciPy \citep{o2021pycham}. The Multi-Scale Infrastructure for Chemistry and Aerosols (MUSICA) is implemented in C++ and provides an option to run on accelerators such as GPUs \citep{acom.software.musica-vision}. For the EMAC model, a source-to-source tool translates the KPP-generated Fortran code into CUDA kernels to solve the chemical kinetics on GPUs \citep{alvanos2017gpu}. TChem-atm couples gas-phase and aerosol chemistry and runs on CPUs and GPUs through the Kokkos performance-portability library, constructing the Jacobian automatically for a range of stiff ODE solvers \citep{diaz2026tchem}. These developments target the efficient forward simulation of process-based atmospheric models, which are built on explicit physical and chemical processes and are therefore interpretable, since the driving forces in the model can be perturbed to test their influence on the output. However, process-based models are limited by the processes that they do not resolve, and amending the model structure or tuning its parameters against observations remains laborious and computationally expensive.

Data-driven approaches, for example neural networks trained on historical atmospheric observations, offer flexible and fast emulation of atmospheric processes with comparable accuracy. ChemNNE is a neural-network emulator of the atmospheric chemical ODE that is considerably faster than the numerical solver \citep{liu2025neural}, and TopoFlow is a topography-aware neural network for high-resolution air-quality prediction \citep{kheder2026topoflow}. Neural networks can also partially solve the estimation of rate coefficients in chemical ODEs \citep{spinode, ino}. However, purely data-driven models often lack interpretability and the ability to predict anomalies or extreme events, and their training is data-intensive and becomes infeasible when observations are sparse. Neural networks are trained by gradient descent, which relies on automatic differentiation, and although this is possible in Fortran, the majority of neural-network models are developed in Julia or in Python frameworks such as PyTorch and JAX.

Hybrid models that combine physically based process models with data-driven neural networks are expected to unite the flexibility of neural networks with the interpretability and physical constraints of process-based models, enabling both fast emulation and learning from data and an understanding of the underlying processes \citep{shen2023differentiable, innes2019differentiable, gelbrecht2023differentiable}. For example, the hybrid weather and climate model NeuralGCM, which combines a differentiable dynamical core with neural networks, both implemented in JAX, achieves a performance comparable to the best machine-learning and process-based models \citep{kochkov2024neural}. In ecohydrology, JAX-CanVeg integrates a land-surface model with neural-network components and improves the simulation \citep{jiang2025jax}. Such hybrid models require the process model itself to be differentiable, so that gradients can be propagated through it to the network parameters. General-purpose differentiable modelling tools for reaction networks exist, such as Catalyst.jl in Julia \citep{loman2023catalyst}, but to the best of our knowledge no tool generates differentiable code from mechanisms written in the KPP language, in which atmospheric mechanisms such as the MCM are distributed and used.

Differentiability matters most for inverse problems, in which model parameters or inputs are inferred from observations of the model output. Estimating rate coefficients from chamber or flow-tube observations is such a problem, and it is often ill-posed because the observations constrain only some combinations of the rate coefficients, so that the identifiability of the parameters should be analysed before they are estimated \citep{guillaume2019introductory}. Another example is the inversion of emissions from concentration observations, which has traditionally relied on variational data assimilation with hand-derived adjoint models \citep{elbern2007emission, henze2007development}. To learn how emissions depend on environmental drivers, the retrieved emission fields can be used to train a neural network offline, in the same way as data assimilation has been combined with machine learning to learn model dynamics from sparse and noisy observations \citep{brajard2020combining}. Such a two-step approach, however, discards the information that the individual retrievals cannot resolve, whereas online training, in which the network is fitted to the observations through the chemical model, requires the gradient of the model output with respect to the network parameters.

To address this, we present KPPax, a kinetic preprocessor implemented in the JAX framework that generates differentiable, just-in-time compiled, and vectorisable chemistry code from mechanisms written in the KPP language, which is directly usable within the data-driven modelling ecosystem. KPPax evaluates the rate law with sparse operations in log space, which keeps it differentiable, and integrates the stiff system with the implicit solvers and the forward- and reverse-mode adjoints of the \texttt{diffrax} library \citep{kidger2022neural}, which enables inverse modelling, parameter estimation, and hybrid modelling of physical and neural models. Section~\ref{sec:impl} describes the implementation. Section~\ref{sec:param} uses the sensitivities from KPPax to analyse the identifiability of the rate coefficients of a benchmark mechanism before estimating them from noisy synthetic observations. Section~\ref{sec:hybrid} trains a neural-network correction of an emission inventory and shows that online fine-tuning through the chemical model improves on the offline two-step approach. Section~\ref{sec:discussion} discusses the limitations, and Sect.~\ref{sec:conclusion} concludes.

\section{Implementation of differentiable kinetic pre-processing}
\label{sec:impl}

\subsection{The chemistry kinetic problem}
An atmospheric chemical reaction system can be mathematically formulated by a coupled system of stiff ordinary differential equations (ODEs). Consider a reaction system of n species and m reactions. We denote the concentration of species $i$ as $y_i$ and reaction $j$ as $r_j$, with $1\leq i \leq n,1\leq j \leq m$. Let $\boldsymbol{y}=[y_1, ..., y_n]^T$ be a vector of concentration of all the species.

The stoichiometric coefficients define the quantitative relationship of the reactants and the products involved in the reaction. We define the stoichiometric coefficient $s_{i,j}^-$ as the number of molecules of species $i$ that react and consumed in the reaction $j$, and similarly $s_{i,j}^+$ as the number of molecules produced. The reaction $j$ can be generically formulated as:
\begin{equation}
    \sum s_{i,j}^-  y_i \xrightarrow{} \sum s_{i,j}^+  y_i
\end{equation}
The kinetic reaction rate law define the reaction rate $\omega_j$ of the reaction $j$ as a proportion with the multiplication of the concentration of the reactants raised by the partial order for reactant $s_{i,j}^-$, with the proportionality constant defined as the rate coefficient (rate constant) $k_j$:
\begin{equation}
\label{eq:rate_law}
    \omega_j(t, \boldsymbol{y})=k_j(t)\prod_{i=1}^{n}y_i^{s_{i,j}^-}
\end{equation}
Depending on the reaction system, the rate coefficient may be either a constant or a time-dependent quantity that varies with environmental conditions. For instance, the photolysis rate of NO$_3$ depends on the actinic flux (light intensity), while the rate of cross-reactions among intermediate peroxy radicals depends on the total concentration of peroxy radicals present in the system. The vector $\boldsymbol{k}$ consists of the rate coefficients $k_j$ from all $j$ reactions.

The concentration of species $i$ change at a rate given by the cumulative effect of consumption and production in all reactions:
\begin{equation}
    \label{eq:stoichiometry}
    \frac{dy_i}{dt} = \sum_{j=1}^{m}(s_{i,j}^+ - s_{i,j}^-)\omega_j(t, \boldsymbol{y})
\end{equation}
We organise the stoichiometric coefficients $s_{i,j}^-$ and $s_{i,j}^+$ in matrices $\boldsymbol{S}^-$ and $\boldsymbol{S}^+$ and the reaction rates in a vector $\boldsymbol{\omega} = [\omega_1, ..., \omega_m]^T$, and let $\boldsymbol{S} = \boldsymbol{S}^+ - \boldsymbol{S}^-$, the rate equation~(\ref{eq:stoichiometry}) can be written as:
\begin{equation}
\label{eq:stoichiometry_matrix}
\frac{d\boldsymbol{y}}{dt} = (\boldsymbol{S}^+ - \boldsymbol{S}^-)\boldsymbol{\omega}(t,\boldsymbol{y}) = \boldsymbol{S}\boldsymbol{\omega}(t,\boldsymbol{y}) = f(t, \boldsymbol{y}),
\end{equation}
This forms an initial value problem and the solution $\boldsymbol{y}(t) \in \mathbb{R}^n$ represents the time evolution of the concentration, starting from the initial concentration $\boldsymbol{y}(t_0)$ over the time span $(t_0, t)$:
\begin{equation}
\label{eq:integrate}
    \boldsymbol{y}(t) = \boldsymbol{y}(t_0) + \int_{t_0}^{t} f\big(\tau, \boldsymbol{y}(\tau)\big)\, d\tau
\end{equation}

\subsection{Reaction parsing}
The proposed KPPax adopts the KPP language as a description syntax for kinetic reaction system. The KPP language, developed by the KPP project, consists of sections, commands, and inline codes. KPP sections contain the description of reaction schemes, including the species and reactions. KPP commands control the code generation and the behaviour of the generated code. More custom behaviour can be implemented as inline code, and be inserted into a certain location in the generated code. KPPax uses only the sections that describe the mechanism, i.e. the species and the equations, because its objective is to generate the chemistry code alone, and it is the responsibility of the user to provide the initial concentrations and to construct the customised simulation.

Following the taxonomy of reaction parsing in \citep{damian2002kinetic}, KPPax parses the kinetic equations in a mixture of the totally integrated approach and the preprocessing approach. The former parses the chemical mechanism once into a set of coefficient arrays that are subsequently evaluated by generic array operations at run time, and the latter generates mechanism-specific source code ahead of time, as KPP itself does. We utilise the parsing module from \texttt{pykpp}, which defines and parse the KPP language using the \texttt{pyparsing} module and returns a list of reactions composed of reactants, products, and rate coefficient expression. The stoichiometric coefficients are converted into sparse array literals in the coordinate list (COO) format, with row/column indices and values stored in separate arrays $(\boldsymbol{i}, \boldsymbol{j}, \boldsymbol{v})$. The time dependent rate coefficients are converted from Fortran expression to JAX expression and wrapped as a python function.

To exploit the computational power of accelerators, we implement the rate law in fully sparse matrix operation. Rather than materialising $\boldsymbol{S}^-$ and $\boldsymbol{S}$ as dense matrices, the rate law (Eq.~\ref{eq:rate_law}-\ref{eq:stoichiometry_matrix}) is evaluated directly from their COO representation using index-gather and segment-reduction primitives, as summarised in Algorithm~\ref{alg:rate_law}. The reactant product term in Eq.~\ref{eq:rate_law} is computed as a segment product: partial powers $y_i^{s_{i,j}^-}$ are gathered at the non-zero row indices of $\boldsymbol{S}^-$ and reduced onto the reaction axis $j$ by scatter-multiplication.

\begin{algorithm}[htbp]
\caption{Sparse evaluation of the rate law}
\label{alg:rate_law}
\begin{algorithmic}[1]
\REQUIRE $\boldsymbol{y}$; $\boldsymbol{k}$; $\boldsymbol{S}^-$: $(\boldsymbol{i}_r, \boldsymbol{j}_r, \boldsymbol{v}_r)$; $\boldsymbol{S}$: $(\boldsymbol{i}_s, \boldsymbol{j}_s, \boldsymbol{v}_s)$
\STATE $\boldsymbol{p} \gets \boldsymbol{y}[\boldsymbol{i}_r]^{\,\boldsymbol{v}_r}$ \COMMENT{gather and raise to partial order}
\STATE $\boldsymbol{\omega} \gets \boldsymbol{k} \odot \mathrm{segment\_prod}(\boldsymbol{p}, \boldsymbol{j}_r)$ \COMMENT{scatter-multiply onto reaction axis}
\STATE $\dot{\boldsymbol{y}} \gets \mathrm{segment\_sum}(\boldsymbol{v}_s \odot \boldsymbol{\omega}[\boldsymbol{j}_s],\, \boldsymbol{i}_s)$ \COMMENT{scatter-add onto species axis}
\RETURN $\dot{\boldsymbol{y}}$
\end{algorithmic}
\end{algorithm}

This power-multiply formulation is not differentiable in practice, because the reactant product is realised as a scatter-multiplication over repeated reaction indices, for which JAX does not implement a differentiation rule. We avoid this implementation limitation by reformulating the rate law in log space,
\begin{equation}
\label{eq:rate_law_log}
\log \omega_j = \log k_j + \sum_{i} s_{i,j}^- \log y_i,
\end{equation}
which replaces the scatter-multiplication with a scatter-\emph{addition}, $\mathrm{segment\_sum}$, before a single exponentiation recovers $\boldsymbol{\omega}$:
\begin{algorithm}[htbp]
\caption{Sparse evaluation of the rate law in log space}
\label{alg:rate_law_log}
\begin{algorithmic}[1]
\REQUIRE $\boldsymbol{y}$; $\boldsymbol{k}$; $\boldsymbol{S}^-$: $(\boldsymbol{i}_r, \boldsymbol{j}_r, \boldsymbol{v}_r)$; $\boldsymbol{S}$: $(\boldsymbol{i}_s, \boldsymbol{j}_s, \boldsymbol{v}_s)$
\STATE $\boldsymbol{q} \gets \boldsymbol{v}_r \odot \log \boldsymbol{y}[\boldsymbol{i}_r]$ \COMMENT{gather and scale log-concentrations by partial order}
\STATE $\boldsymbol{l} \gets \mathrm{segment\_sum}(\boldsymbol{q}, \boldsymbol{j}_r)$ \COMMENT{scatter-add onto reaction axis, i.e. $\log\prod_i y_i^{s_{i,j}^-}$}
\STATE $\boldsymbol{\omega} \gets \boldsymbol{k} \odot \exp(\boldsymbol{l})$ \COMMENT{recover reaction rate}
\STATE $\dot{\boldsymbol{y}} \gets \mathrm{segment\_sum}(\boldsymbol{v}_s \odot \boldsymbol{\omega}[\boldsymbol{j}_s],\, \boldsymbol{i}_s)$ \COMMENT{scatter-add onto species axis}
\RETURN $\dot{\boldsymbol{y}}$
\end{algorithmic}
\end{algorithm}

Since JAX defines a standard gradient rule for $\mathrm{segment\_sum}$, this substitution leaves the sparse indexing pattern and computational cost of Algorithm~\ref{alg:rate_law} unchanged while restoring full differentiability of the rate law with respect to both $y$ and $k$, at the cost of requiring $y$ to remain strictly positive throughout the integration.

Atmospheric chemical mechanisms are numerically stiff, because reactive radical species relax on timescales of milliseconds to seconds, while longer-lived species evolve over hours, so the system's Jacobian eigenvalues span many orders of magnitude. Explicit integration schemes are constrained by this stiffness to prohibitively small time steps for stability, rather than accuracy, and are therefore unsuitable, which is why KPP itself provides implicit and Rosenbrock-type solvers \citep{sandu2003direct}. We integrate Eq.~(\ref{eq:stoichiometry_matrix}) using \texttt{diffrax} \citep{kidger2022neural}, a JAX-native library of differentiable ODE solvers, selecting the implicit Kvaerno family of solvers, which combine embedded error estimates with an implicit treatment of the stiff terms via a diagonally implicit Runge-Kutta scheme, avoiding the step-size collapse of explicit methods. Step size is controlled adaptively by a proportional-integral (PID) controller, which adjusts $\mathrm{d}t$ to meet a prescribed local error tolerance rather than relying on a fixed or manually tuned step size. Because concentrations must remain strictly positive for the log-space rate law (Eq.~\ref{eq:rate_law_log}), we additionally clip $y$ to a small positive floor at each evaluation of the right-hand side, preventing the solver from probing negative or exactly zero concentrations during error-controlled step adjustment. We perform all stiff ODE integration in 64-bit double precision.

\subsection{Automatic differentiation}
Automatic differentiation (AD) computes exact derivatives of a numerical program by decomposing it into elementary operations and applying the chain rule systematically, either forward alongside the primal computation (forward mode) or backward from the output (reverse mode). Once the rate law and its numerical integration are expressed as JAX-traceable array operations, the sensitivity of an output quantity, for instance an observed concentration at time $t_F$, with respect to an input quantity, for instance an emission rate or a rate coefficient, can therefore be obtained automatically, without deriving the corresponding tangent-linear or adjoint equations by hand.

Differentiating through an ODE solve is nonetheless a distinct problem from differentiating a single expression, since the numerical solution is produced by an iterative sequence of solver steps rather than a closed-form function. Two broad strategies exist for obtaining the required sensitivities, namely the \emph{optimize-then-discretize} adjoint, which derives an adjoint ODE from the continuous forward equations and integrates it backwards independently of how the forward solve was discretised, and the \emph{discretize-then-optimize} adjoint, which differentiates the forward numerical scheme itself, step by step, yielding the exact gradient of the numerical solution rather than an approximation to the gradient of the underlying continuous problem \citep{sandu2003direct}. KPP implements both, but its discretize-then-optimize adjoint must be derived analytically for each integrator in turn. For example, \citet{sandu2003direct} derive it explicitly for the linearly implicit Euler method, requiring not only the Jacobian but also its derivative with respect to concentration (the Hessian tensor) and dedicated sparse linear-algebra routines to evaluate it efficiently. They note this construction to be considerably more demanding than their optimize-then-discretize adjoint, and correspondingly, KPP provides hand-derived discretize-then-optimize adjoints for only a subset of its integrators.

\texttt{diffrax} \citep{kidger2022neural} instead obtains a discretize-then-optimize adjoint of the forward solve through reverse-mode AD applied directly to the solver's internal steps, exposing the differentiation strategy as a configurable component of the solve rather than a property fixed by the choice of integrator. Naively differentiating through every accepted step this way would require storing each intermediate solver state, incurring memory cost proportional to the number of steps. The recursive checkpoint adjoint avoids this by checkpointing only a logarithmic subset of states and recomputing the remainder on demand during the backward pass, keeping memory cost sub-linear in step count while still returning the exact discrete gradient. Because this adjoint is generated automatically for any solver JAX can trace, KPPax removes the need to derive a Hessian or a solver-specific adjoint step by hand for every integrator, and with it the correctness risk that \citet{sandu2003direct} themselves highlight for hand-written or general-purpose adjoint-compiler code. The trade-off is computational rather than conceptual: checkpointed recomputation adds cost that a hand-derived discretize-then-optimize adjoint, once implemented, would not incur, particularly for very large mechanisms. We consider this an acceptable exchange for the substantial reduction in implementation and maintenance effort it affords when extending KPPax to new mechanisms or integrators.
\section{Application I: identifiability and parameter estimation}
\label{sec:param}

Rate coefficients of atmospheric chemical mechanisms are uncertain, and many of them can only be constrained indirectly, through laboratory or chamber observations of a few species. A fit of all rate coefficients to such observations is often non-unique, because different combinations of parameter values reproduce the observations equally well. A parameter is non-identifiable when its value cannot be determined uniquely from the data, and this can arise from the structure of the model, the observed quantities and the objective function (structural non-identifiability), or from the experimental design and the observation error (practical non-identifiability) \citep{guillaume2019introductory}. Therefore identifiability is recomended to be assessed routinely before parameters are estimated, so that the estimation is restricted to what the data can support. Derivative-based identifiability analysis requires the sensitivity of the model output with respect to every parameter, which is laborious to derive by hand for a chemical mechanism and inaccurate and expensive to approximate by finite differences. With KPPax, the exact sensitivities are obtained by automatic differentiation through the stiff solver at a cost comparable to a few forward integrations. In this section, we use these sensitivities first to predict, before any fit, which rate coefficients a given experimental design can constrain, and then to estimate the rate coefficients from noisy synthetic observations, which tests these predictions.

\subsection{Method}
\label{sec:param_method}

We parameterise the rate coefficients as $\boldsymbol{k} = \boldsymbol{k}^{\mathrm{ref}} \odot \exp(\boldsymbol{\theta})$, where $\boldsymbol{k}^{\mathrm{ref}}$ are the reference values of the mechanism and $\boldsymbol{\theta}$ is the vector of log correction factors. The observation operator $\boldsymbol{h}(\boldsymbol{\theta})$ stacks the logarithms of the simulated concentrations of all observed species at all observation times and in all experiments. Given observations $\tilde{\boldsymbol{y}}$ with independent Gaussian errors of standard deviation $\sigma$ in log space, the residuals and the least-squares cost are
\begin{equation}
    \label{eq:residual_param}
    \boldsymbol{r}(\boldsymbol{\theta}) = \frac{\boldsymbol{h}(\boldsymbol{\theta}) - \tilde{\boldsymbol{y}}}{\sigma},
    \qquad
    \chi^2(\boldsymbol{\theta}) = \boldsymbol{r}(\boldsymbol{\theta})^{\mathsf{T}} \boldsymbol{r}(\boldsymbol{\theta}).
\end{equation}
Because $\boldsymbol{h}$ and $\boldsymbol{\theta}$ are both logarithmic, the Jacobian $\boldsymbol{J} = \partial \boldsymbol{h} / \partial \boldsymbol{\theta}$ contains the relative sensitivities $\partial \ln y / \partial \ln k$, which are comparable across species and reactions whose magnitudes differ by many orders. Given the number of $\boldsymbol{\theta}$ and $\boldsymbol{y}$, we compute $\boldsymbol{J}$ by forward-mode automatic differentiation through the solver, which propagates one tangent per parameter.

The local practical identifiability at the true parameters is characterised by the Fisher information matrix (FIM)
\begin{equation}
    \label{eq:fim}
    \boldsymbol{F} = \frac{1}{\sigma^2} \boldsymbol{J}^{\mathsf{T}} \boldsymbol{J} = \boldsymbol{V} \boldsymbol{\Lambda} \boldsymbol{V}^{\mathsf{T}},
\end{equation}
which for noise-free observations equals the Gauss--Newton approximation of the Hessian of $\chi^2/2$ at the truth. Each eigenvector $\boldsymbol{v}_m$ (a column of $\boldsymbol{V}$) is a direction in parameter space, i.e. a combination of rate coefficients, and its eigenvalue $\lambda_m$ measures how strongly the data constrain this combination. The Cram\'er--Rao bound gives the smallest covariance $\boldsymbol{F}^{-1}$ that any unbiased estimator can attain, which corresponds to a standard deviation of $\lambda_m^{-1/2}$ along $\boldsymbol{v}_m$ and to a standard deviation of
\begin{equation}
    \label{eq:cr_sd}
    \sigma^{\mathrm{CR}}_p = \Big[\big(\boldsymbol{F}^{-1}\big)_{pp}\Big]^{1/2} = \Big(\sum_m \frac{V_{pm}^2}{\lambda_m}\Big)^{1/2}
\end{equation}
for the individual parameter $\theta_p$. We classify a rate coefficient as identifiable if $\sigma^{\mathrm{CR}}_p < 0.1$, i.e. if its one-standard-deviation uncertainty is about 10\% (a factor of $e^{0.1} \approx 1.1$), which is small compared with the factor-of-two uncertainty ($\ln 2 \approx 0.69$) of the initial guesses used below. In the experiment of Sect.~\ref{sec:param_design}, the classification does not depend on the exact threshold, because the values of $\sigma^{\mathrm{CR}}_p$ jump from 0.04 to 0.15 between the identifiable and the remaining parameters (Sect.~\ref{sec:param_results}). Directions with eigenvalues close to zero are combinations of rate coefficients that the observations cannot distinguish, and their loadings $V_{pm}$ show which reactions are confounded. This analysis requires only the Jacobian at a single parameter point and is performed before any fit.

To test the predictions, we estimate the parameters from noisy synthetic observations. In each of 20 trials, we draw a new noise realisation of $\tilde{\boldsymbol{y}}$ at the truth and a new initial guess $\boldsymbol{\theta}^{(0)} \sim \mathcal{N}(0, (\ln 2)^2 \boldsymbol{I})$, i.e. initial rate coefficients that typically deviate from the truth by a factor of 2, and minimise $\chi^2$ with the Levenberg--Marquardt algorithm, i.e. Gauss--Newton iterations with adaptive damping, of the \texttt{optimistix} library \citep{rader2024optimistix}, using the same forward-mode Jacobian and at most 200 iterations. All trials are solved together under jax native vectorisation \texttt{jax.vmap}. For each parameter, we compare the root-mean-square (RMS) deviation of the estimates from the truth across the trials with $\sigma^{\mathrm{CR}}_p$. For identifiable parameters, the two should agree. For non-identifiable parameters, the estimates are expected to remain scattered at least as widely as the initial guesses, because the data cannot pull them back, even though the fit to the observations is as good as at the truth.

\subsection{Experimental design}
\label{sec:param_design}

We use the POLLU scheme \citep{verwer1994gauss} with its reference rate coefficients, which are constant in time, and estimate all 25 rate coefficients. The design mimics a flow-tube experiment, in which the reaction time is fixed and the injected precursor concentrations are varied. POLLU initialises six species (NO, O$_3$, HCHO, CO, ALD, and SO$_2$), and we generate eight initial conditions by multiplying each of these initial concentrations by an independent log-normal factor with a log-space standard deviation of $\ln 2$. All other species start from zero. Each initial condition is integrated for 60~min, and all 20 species are observed once at the end, which gives 160 observations with a log-space noise of $\sigma = 0.05$. The ODEs are integrated with the Kvaerno3 solver with relative and absolute tolerances of $10^{-8}$ and $10^{-9}$~ppm.

If all species were observed continuously in time, all 25 rate coefficients of POLLU would be globally structurally identifiable, as we verified with the \texttt{StructuralIdentifiability.jl} package \citep{dong2023differential}. Any non-identifiability in this experiment is therefore practical and arises from the design, which observes a single time after a fixed reaction time. Such a design is outside the scope of the structural analysis, which assumes continuous-time outputs, whereas the FIM applies to any set of observations.

\subsection{Results}
\label{sec:param_results}

The automatically differentiated Jacobian ($160 \times 25$) agrees with the central finite-difference approximation to a relative error of $8.7 \times 10^{-7}$, and it takes 0.77~s, compared with 28.7~s for the 50 evaluations of the forward model required by the finite differences. The finite-difference approximation also depends on the choice of the step size, whereas automatic differentiation gives the exact derivative of the numerical solution.

\begin{figure}
    \centering
    \begin{minipage}[c]{0.44\linewidth}
        \textbf{(a)}\\
        \includegraphics[width=\linewidth]{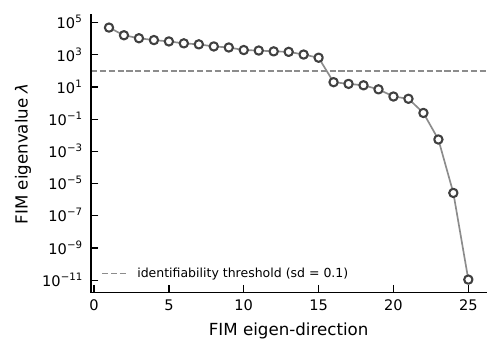}\\[0.5em]
        \textbf{(b)}\\
        \includegraphics[width=\linewidth]{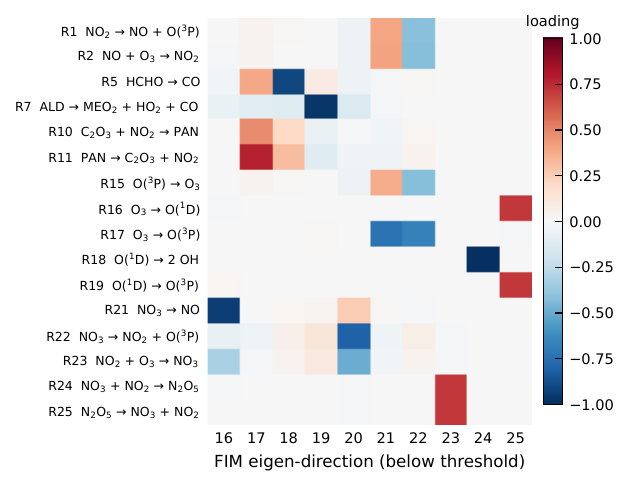}
    \end{minipage}\hfill
    \begin{minipage}[c]{0.54\linewidth}
        \textbf{(c)}\\
        \includegraphics[width=\linewidth]{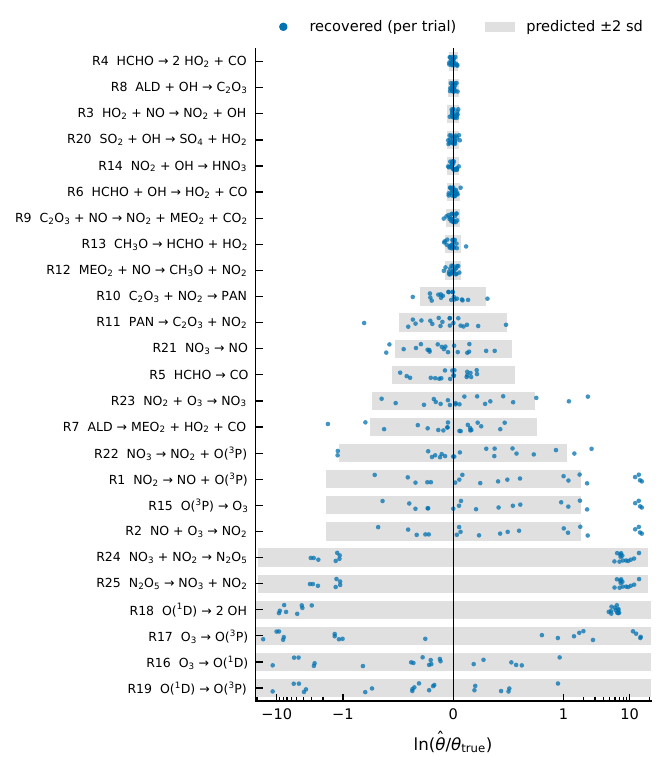}
    \end{minipage}
    \caption{Identifiability analysis and estimation of the 25 POLLU rate coefficients. (a) Eigenvalues of the Fisher information matrix in descending order, where the dashed line marks $\lambda = 100$, which corresponds to a Cram\'er--Rao standard deviation of 0.1 along the eigenvector. (b) Loadings of the ten eigenvectors below this threshold on the rate coefficients that contribute to them. (c) Estimated log rate coefficients relative to the truth in the 20 trials (dots), compared with the interval of $\pm 2 \sigma^{\mathrm{CR}}_p$ predicted from the FIM before the fit (grey bars). The horizontal axis is linear between $-1$ and 1 and logarithmic outside. Reactions are numbered as in POLLU.}
    \label{fig:rate}
\end{figure}

The eigenvalues of the FIM span 16 orders of magnitude (Fig.~\ref{fig:rate}a). Fifteen eigen-directions are constrained to a standard deviation better than 0.1, and the remaining ten fall off steeply, with the four smallest eigenvalues below 1 and the smallest at $10^{-11}$. Since the well-constrained directions are combinations of rate coefficients, only nine individual rate coefficients have $\sigma^{\mathrm{CR}}_p < 0.1$ (Fig.~\ref{fig:rate}c). These are the reactions of the HO$_x$, HCHO, ALD, and SO$_2$ chemistry (R3, R4, R6, R8, R9, R12, R13, R14, and R20). The loadings of the weakly constrained eigenvectors (Fig.~\ref{fig:rate}b) group the non-identifiable rate coefficients into three clusters of fast reactions. These are the NO--NO$_2$--O$_3$ photostationary cycle (R1, R2, and R15), the equilibrium between NO$_3$, NO$_2$, and N$_2$O$_5$ (R24 and R25), and the fast interconversion of O$_3$, O($^1$D), and O($^3$P) (R16--R19). Within each cluster, the reactions are fast compared with the reaction time of 60~min, so the observed concentrations are set by the ratios of the rate coefficients, and their absolute values are not constrained. For example, the eigenvector that loads equally on R24 and R25 corresponds to a simultaneous increase of both rate coefficients, which leaves the N$_2$O$_5$ equilibrium constant $k_{24}/k_{25}$ unchanged.
The noisy estimation confirms these predictions (Fig.~\ref{fig:rate}c). For the nine identifiable rate coefficients, the RMS deviation of the estimates from the truth agrees with $\sigma^{\mathrm{CR}}_p$ to within about 30\%, which shows that the local, linearised prediction from a single Jacobian quantifies the achievable accuracy correctly. The seven rate coefficients with intermediate $\sigma^{\mathrm{CR}}_p$ of 0.15--0.57 are estimated with errors of the predicted size, which is somewhat larger for R22 and R23. The non-identifiable rate coefficients drift far from the truth, by up to 15 in log units, i.e. by more than six orders of magnitude. For the cluster R1, R2, and R15, the RMS deviation of 6.3 is seven times the Cram\'er--Rao prediction, because the drift follows a curved valley of the cost function along which the linearisation at the truth is not valid. For the O$_3$--O($^1$D)--O($^3$P) cluster, the RMS deviation is much smaller than the Cram\'er--Rao value, which in this case only indicates that the cost is flat, so that the spread of the estimates is set by the initial guesses and the finite number of iterations. The drift is not random, because the estimates within a cluster move together and keep their ratios constant, which is visible in Fig.~\ref{fig:rate}c as identical estimates of R1, R2, and R15, and of R24 and R25, in each trial. The FIM therefore identifies correctly which combinations of rate coefficients the design cannot constrain, but it underestimates how far the estimates can move along these combinations.
All fits reach the noise level of the observations. The median of $\chi^2/n$ over the trials is 0.85 at the estimates and 1.01 at the truth, and the reduction is consistent with the number of parameters $p$ that the fit can adjust to the noise, since $1 - p/n$ is 0.84--0.91 for $p$ between the 15 well-constrained directions and all 25 parameters. Only two of the 20 trials met the convergence tolerance of the Levenberg--Marquardt algorithm, whereas the others reached the maximum of 200 iterations. These trials did not fail, but kept creeping along the flat valleys of the non-identifiable clusters, where each step reduces the cost only marginally, while their estimates of the identifiable rate coefficients agree with the Cram\'er--Rao prediction (Fig.~\ref{fig:rate}c). This behaviour is itself a symptom of non-identifiability, and it illustrates why the analysis should precede the fit: once the FIM has revealed the confounded clusters, their rate coefficients can be fixed at literature values or reparameterised by their ratios, and only the identifiable combinations are estimated.

\section{Application II: hybrid neural emission inversion}
\label{sec:hybrid}

Emission inventories are a major source of uncertainty in chemical transport modelling. Their dependence on environmental drivers, such as temperature, light, or population density, is usually prescribed by simple empirical functions. Concentration observations constrain emissions only indirectly, through the chemistry, so a correction to the inventory has to be inferred by inverting the chemical model, as in four-dimensional variational (4D-Var) emission inversion \citep{elbern2007emission}. A neural network that maps environmental drivers to an emission correction can then be trained offline, by regressing it on the emission fields retrieved by such an inversion. However, the accuracy of the network is then limited by the retrieved fields themselves, because each retrieval is regularised towards the prior inventory and is constrained only by the observations of its own scenario, so that information that the observations do contain, but that a single retrieval cannot isolate from the prior, is lost before the network is trained. Here, we use KPPax to learn a neural-network correction of the emission inventory as a function of its environmental drivers. Because the network sits inside the kinetic ODE, its training requires gradients of the simulated concentrations with respect to the network parameters, propagated through the stiff chemistry, which is exactly what the differentiable solver provides. We compare the offline training with online training through the ODE in a synthetic experiment where the true correction is known.

\subsection{Experimental setup}
\label{sec:hybrid_setup}

We use the POLLU scheme \citep{verwer1994gauss} (20 species, 25 reactions) in a 0-dimensional box model with emission and dilution towards a background concentration $\boldsymbol{c}_{\mathrm{bg}}$:
\begin{equation}
    \label{eq:box}
    \frac{\mathrm{d}\boldsymbol{y}}{\mathrm{d}t} = \boldsymbol{S}\,\boldsymbol{\omega}\big(t, \boldsymbol{y}\big) + \boldsymbol{E}(t) - \frac{\boldsymbol{y} - \boldsymbol{c}_{\mathrm{bg}}}{\tau},
\end{equation}
where the first term is the rate law of Eq.~(\ref{eq:stoichiometry_matrix}) generated by KPPax, and $\tau = 12$~h is the dilution time scale. The rate coefficients of POLLU are constant, so we impose a diurnal cycle on its eight photolysis reactions by scaling them with a smooth daylight factor that equals 1 at noon and vanishes between 18:00 and 06:00. Concentrations are in ppm and time is in minutes, following POLLU.

\begin{figure}
    \centering
    \includegraphics[width=\linewidth]{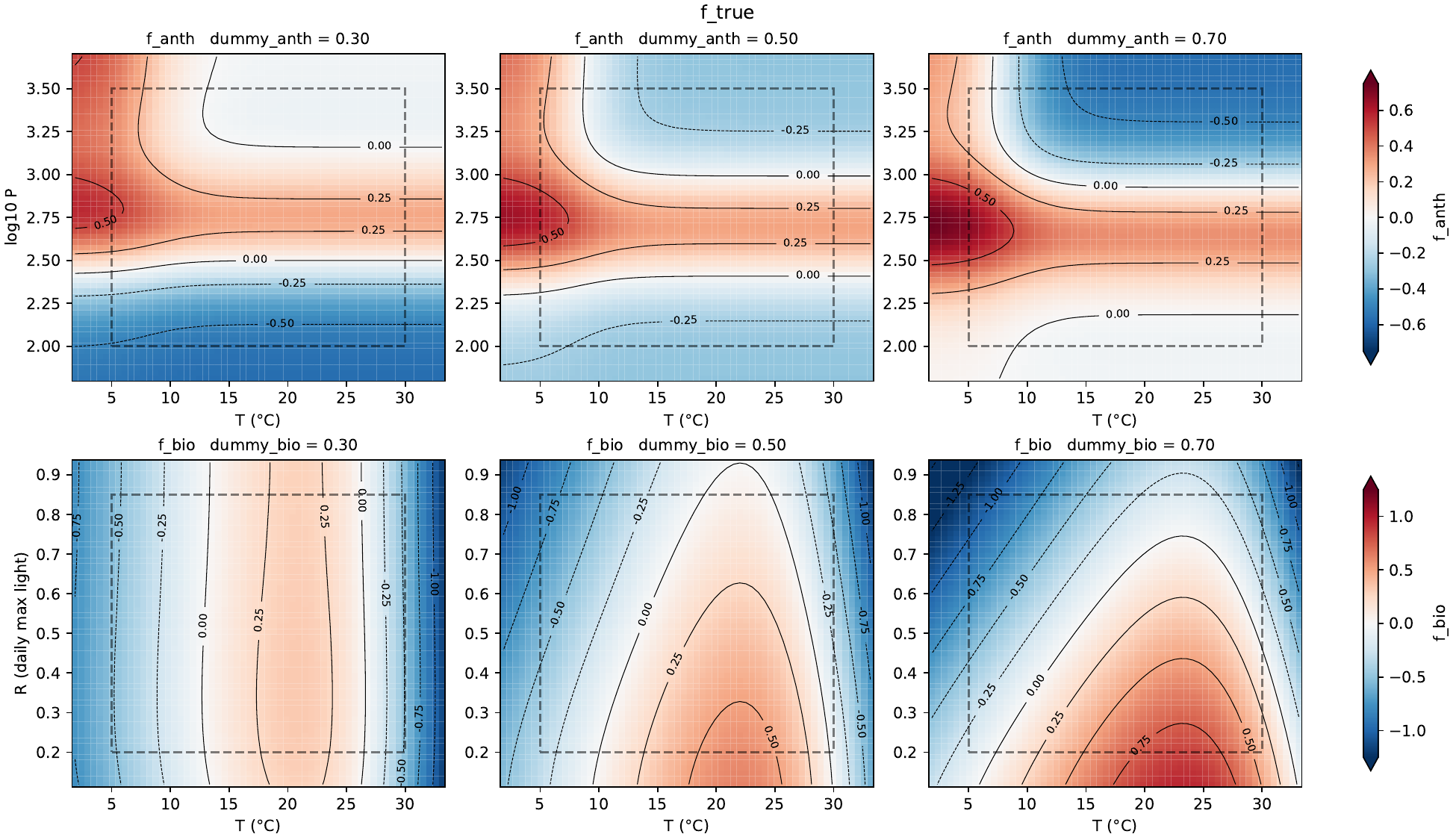}
    \caption{True log emission correction $f_s^{\mathrm{true}}$ (Eqs.~\ref{eq:f_true_anth} and~\ref{eq:f_true_bio}) for the anthropogenic (top) and biogenic (bottom) sectors, at three values of the synthetic covariate (columns). The dashed boxes mark the training ranges, and the regions outside them are the extrapolation bands.}
    \label{fig:f_true}
\end{figure}

Emissions come from an anthropogenic sector ($s = \mathrm{anth}$), which emits NO, NO$_2$, CO, SO$_2$, HCHO, and ALD, and a biogenic sector ($s = \mathrm{bio}$), which emits ALD and HCHO. The total emission rate is
\begin{equation}
    \label{eq:emission}
    \boldsymbol{E}(t) = \sum_{s} \boldsymbol{e}_s \, g_s\big(\boldsymbol{x}(t)\big) \, p_s(t) \, \exp\big[f_s\big(\boldsymbol{x}(t)\big)\big].
\end{equation}
Here, $\boldsymbol{e}_s$ is a fixed speciated base emission rate, and $p_s(t)$ is a diurnal profile known a priori and normalised to a daily mean of one, which follows a traffic double peak at 08:00 and 18:00 for the anthropogenic sector and the daylight factor for the biogenic sector. The covariate vector $\boldsymbol{x} = (T, R, \log_{10}P, d_{\mathrm{anth}}, d_{\mathrm{bio}})$ consists of the temperature $T$ (in $^\circ$C), the daily maximum light level $R \in [0, 1]$, the population density $P$, and two synthetic covariates, which can be interpreted as the share of fossil-fuel heating ($d_{\mathrm{anth}}$) and the vegetation type or thermal acclimation ($d_{\mathrm{bio}}$). The inventory scaling $g_s$ is a deliberately crude function of the covariates, in which anthropogenic emissions scale linearly with population density, whereas biogenic emissions follow a fixed exponential temperature response and depend linearly on light:
\begin{equation}
    \label{eq:inventory}
    g_{\mathrm{anth}} = 10^{\log_{10}P - 3}, \qquad
    g_{\mathrm{bio}} = \exp\big[0.09\,(T - 30)\big]\, R.
\end{equation}
The log correction factor $f_s$ is the unknown to be learned, and the inventory corresponds to $f_s = 0$. The true correction depends on three covariates per sector and is constructed to be smooth and nonlinear, so that neither a linear nor an additive model can reproduce it:
\begin{align}
    \label{eq:f_true_anth}
    f_{\mathrm{anth}}^{\mathrm{true}} &= 0.6\, e^{-\left[(\log_{10}P - 2.7)/0.35\right]^2}
        + (0.3 + d_{\mathrm{anth}})\, \sigma\!\left(\tfrac{8 - T}{2}\right) \sigma\big(3\,(\log_{10}P - 2.8)\big) \nonumber \\
        &\quad + 0.3 \sin(2\pi d_{\mathrm{anth}}) \tanh\big(2\,(\log_{10}P - 2.75)\big) - 0.3, \\
    \label{eq:f_true_bio}
    f_{\mathrm{bio}}^{\mathrm{true}} &= 0.07\,(T - T_{\mathrm{opt}}) - \ln\big(1 + e^{0.3\,(T - T_{\mathrm{opt}})}\big)
        + \ln\frac{2}{\sqrt{1 + 3R^2}}
        + 0.8 \sin(2\pi d_{\mathrm{bio}})\,(R - 0.55) + 0.49,
\end{align}
where $\sigma$ is the logistic function and $T_{\mathrm{opt}} = 23 + 6\, d_{\mathrm{bio}}$. The anthropogenic correction (Eq.~\ref{eq:f_true_anth}) contains a non-monotonic dependence on population density, peaking in suburban densities, and a cold-day heating term in populated areas that grows with $d_{\mathrm{anth}}$. Combined with the inventory, the biogenic correction (Eq.~\ref{eq:f_true_bio}) gives a temperature response that rises at 0.16~$^\circ$C$^{-1}$ at low temperature and has an optimum near $T_{\mathrm{opt}}$, and a light dependence $2R/\sqrt{1 + 3R^2}$ that saturates instead of growing linearly. Each sector also includes a sinusoidal interaction with its synthetic covariate, and the constants centre both corrections at approximately zero over the training ranges (Fig.~\ref{fig:f_true}). Since biogenic emissions are small at low temperature, $f_{\mathrm{bio}}$ there has little effect on any concentration and is only weakly constrained by the observations.

In each scenario, the covariates are sampled uniformly from their training ranges. The daily mean temperature is chosen such that the diurnal temperature cycle, which has a range of 4--14~$^\circ$C and a maximum between 13:00 and 16:00, stays within 5--30~$^\circ$C, while $R$, $\log_{10}P$, and $d_{\mathrm{anth}}, d_{\mathrm{bio}}$ are drawn from $[0.2, 0.85]$, $[2, 3.5]$, and $[0.1, 0.9]$, respectively. Background O$_3$ and CO are also sampled per scenario. Temperature varies over the day, so $\boldsymbol{x}$ and hence $f_s$ are functions of time. We generate 150 training, 30 validation, and 30 interpolation test scenarios. For each covariate, we additionally generate 20 extrapolation test scenarios in which that covariate lies in bands extending its training range by 12.5\% of the range width on either side. Each scenario is spun up for one day with the true emissions, and then integrated over a one-day window from midnight, and the post-spin-up state is assumed to be known exactly and is used, without noise, as the initial state of the assimilation window. Observations are taken hourly from 01:00 to 24:00 and perturbed with multiplicative log-normal noise with a log-space standard deviation of 0.05. We consider two observation sets, namely all 20 species and only the five species routinely measured at air-quality monitoring stations (O$_3$, NO, NO$_2$, CO, and SO$_2$). Each set is used with and without noise, giving four settings.

\subsection{Neural emission correction}
\label{sec:hybrid_nn}

The unknown correction $f_s$ is represented by one multilayer perceptron (MLP) per sector, which takes the three covariates of that sector as input, scaled linearly from their training ranges to $[-1, 1]$. Each MLP has two hidden layers of width 32 with softplus activations, and its output is bounded as $f_s = f_{\max} \tanh(\mathrm{MLP}_s / f_{\max})$ with $f_{\max} = 3$, so that an untrained network cannot push the chemistry into extreme states. The weights and bias of the last layer are initialised to zero, so the network reproduces the inventory ($f_s = 0$) at the start of training. The two MLPs together have 2434 parameters, which we denote by $\boldsymbol{\theta}$.

The correction varies in time through the temperature, and we evaluate it at hourly knots $t_j$ ($j = 0, \dots, 24$) over the one-day window, as $f_{s,j} = f_s\big(\boldsymbol{x}(t_j); \boldsymbol{\theta}\big)$, and interpolate between the knots with a cubic Hermite spline. This gives the same time representation of the correction as the 4D-Var control variable in Sect.~\ref{sec:hybrid_offline}, so that the offline and online approaches differ only in how the knot values are constrained. Both approaches measure the misfit to the observations $\boldsymbol{y}^{\mathrm{obs}}$ in log space, as in Eq.~(\ref{eq:residual_param}) but with a detection limit:
\begin{equation}
    \label{eq:residual}
    r_{t,i} = \frac{\ln \max(y_{t,i}, y_{\mathrm{DL}}) - \ln \max(y^{\mathrm{obs}}_{t,i}, y_{\mathrm{DL}})}{\sigma_{\mathrm{obs}}},
\end{equation}
for every observation time $t$ and observed species $i$, where $y_{\mathrm{DL}} = 5 \times 10^{-5}$~ppm is a detection limit that prevents the logarithm from amplifying differences between near-zero concentrations, and $\sigma_{\mathrm{obs}} = 0.05$ in all settings. All ODEs are integrated with the Kvaerno3 solver with relative and absolute tolerances of $10^{-6}$ and $10^{-12}$~ppm.

\subsection{Offline two-step training}
\label{sec:hybrid_offline}

The offline approach first retrieves the correction separately for each scenario $n$ by 4D-Var and then trains the network to reproduce the retrievals. The control variable of scenario $n$ is the vector $\boldsymbol{\delta}_n$ of knot values of both sectors, and the retrieval minimises
\begin{equation}
    \label{eq:var4d}
    J_n(\boldsymbol{\delta}_n) = \frac{1}{2} \sum_{t,i} r_{t,i}^2 + \frac{1}{2} \sum_s \boldsymbol{\delta}_{n,s}^{\mathsf{T}} \boldsymbol{B}^{-1} \boldsymbol{\delta}_{n,s},
    \qquad B_{jl} = \sigma_b^2 \exp\left(-\frac{|t_j - t_l|}{L}\right),
\end{equation}
where the background term keeps the retrieval close to the inventory ($\boldsymbol{\delta} = 0$) with a prior standard deviation $\sigma_b$ and a temporal correlation length $L$. We minimise Eq.~(\ref{eq:var4d}) by Gauss--Newton iterations with Levenberg--Marquardt damping for 15 iterations \citep{courtier1994strategy}. The Jacobian of the residuals with respect to the 50 control variables is obtained by forward-mode automatic differentiation through the stiff solver because the number of controls is much smaller than the number of residuals. The posterior standard deviation $\sigma^{\mathrm{post}}_{n,j,s}$ of each knot value is taken from the diagonal of the inverse Gauss--Newton Hessian $\boldsymbol{H}_n = \boldsymbol{J}_n^{\mathsf{T}} \boldsymbol{J}_n + \boldsymbol{B}^{-1}$ at the minimum.

The prior hyperparameters $(\sigma_b, L)$ are chosen from the data alone rather than from the truth. For each candidate pair, we sum the Laplace approximation of the log marginal likelihood,
\begin{equation}
    \label{eq:evidence}
    \ln p(\boldsymbol{y}^{\mathrm{obs}}_n \mid \boldsymbol{B}) \approx -J_n(\boldsymbol{\delta}_n^*) + \frac{1}{2} \ln\det \boldsymbol{B}^{-1} - \frac{1}{2} \ln\det \boldsymbol{H}_n + \mathrm{const},
\end{equation}
over the training scenarios and select the maximum on a grid. Unlike the cost $J_n$ itself, the evidence penalises prior freedom that the data cannot use, and it therefore discriminates between priors that fit the observations equally well. The selected prior has $\sigma_b = 0.35$ in three of the four settings and $\sigma_b = 0.25$ with noise-free station observations. The selected correlation lengths are 48~h and 96~h with all species observed, and 192~h, the largest value on the grid, with station observations, which means that these observations cannot resolve the variation of the correction within a day.

In the second step, the network is trained to reproduce the retrieved knot values $\hat{\boldsymbol{\delta}}$ from the covariates at the knots of all training scenarios, by minimising the weighted mean squared error
\begin{equation}
    \label{eq:offline}
    \mathcal{L}_{\mathrm{off}}(\boldsymbol{\theta}) = \frac{1}{N} \sum_{n,j,s} w_{n,j,s} \Big[f_s\big(\boldsymbol{x}_n(t_j); \boldsymbol{\theta}\big) - \hat{\delta}_{n,j,s}\Big]^2,
    \qquad w_{n,j,s} \propto \big(\sigma^{\mathrm{post}}_{n,j,s}\big)^{-2},
\end{equation}
where $N$ is the number of terms and the weights are normalised to a mean of one per sector. Without weighting, retrievals that remain at the prior because the observations carry no information, such as biogenic emissions at night, would enter the regression as if they were genuine values of zero. We train with the Adam optimiser \citep{kingma2015adam} for 5000 full-batch steps with a learning rate decaying from $10^{-3}$ along a cosine schedule. No ODE is solved in this step. The weighted regression was more accurate than the unweighted one for the anthropogenic sector in all settings and for the biogenic sector when all species are observed, whereas the two performed similarly for the biogenic sector with station observations. We therefore use the weighted regression as the offline reference below.

\subsection{Online fine-tuning through the ODE}
\label{sec:hybrid_online}

In the online approach, the network parameters are fitted directly to the concentration observations of all training scenarios, with the network inside the ODE (Eqs.~\ref{eq:box} and~\ref{eq:emission}), by minimising
\begin{equation}
    \label{eq:online}
    \mathcal{J}(\boldsymbol{\theta}) = \frac{1}{2} \sum_{n \in \mathrm{train}} \sum_{t,i} r_{n,t,i}\big(\boldsymbol{\delta}_n(\boldsymbol{\theta})\big)^2,
    \qquad \boldsymbol{\delta}_n(\boldsymbol{\theta}) = \Big[f_s\big(\boldsymbol{x}_n(t_j); \boldsymbol{\theta}\big)\Big]_{j,s}.
\end{equation}
The data term is the same as in Eq.~(\ref{eq:var4d}), but the knot values of every scenario are now produced by the shared network rather than being free control variables, and no background term is used, because the shared network already restricts the knot values to a smooth function of the covariates. This allows information from all scenarios to be combined, including information that no single scenario constrains well enough to be retrieved on its own. Training requires the gradient of the misfit with respect to $\boldsymbol{\theta}$, which passes through the stiff chemistry and is therefore only available with a differentiable ODE such as the one generated by KPPax.

We initialise the network with the offline solution and fine-tune it by minimising Eq.~(\ref{eq:online}) with Gauss--Newton iterations with Levenberg--Marquardt damping. By the chain rule, the Jacobian of the residuals of scenario $n$ with respect to the parameters factorises as $\boldsymbol{J}_{\delta,n} \boldsymbol{D}_n$, where $\boldsymbol{J}_{\delta,n}$ is the residual Jacobian with respect to the 50 knot values used in 4D-Var, and $\boldsymbol{D}_n = \partial \boldsymbol{\delta}_n / \partial \boldsymbol{\theta}$ is the Jacobian of the network at the knots, which requires no ODE solve. The Gauss--Newton Hessian and gradient are accumulated over the training scenarios as
\begin{equation}
    \label{eq:gn}
    \boldsymbol{H} = \sum_n \boldsymbol{D}_n^{\mathsf{T}} \boldsymbol{J}_{\delta,n}^{\mathsf{T}} \boldsymbol{J}_{\delta,n} \boldsymbol{D}_n,
    \qquad \boldsymbol{g} = \sum_n \boldsymbol{D}_n^{\mathsf{T}} \boldsymbol{J}_{\delta,n}^{\mathsf{T}} \boldsymbol{r}_n,
    \qquad \Delta\boldsymbol{\theta} = -\big[\boldsymbol{H} + \lambda \operatorname{diag}(\boldsymbol{H})\big]^{-1} \boldsymbol{g},
\end{equation}
and the damping $\lambda$, initialised to $10^{-2}$, is divided by 3 after an accepted step and multiplied by 4 after a rejected one. The ODE sensitivities $\boldsymbol{J}_{\delta,n}$ are therefore the same as in one 4D-Var iteration, and their cost does not grow with the number of network parameters. The parameters only determine the size of $\boldsymbol{H}$, which has one row and one column per parameter ($2434 \times 2434$), and the damped system in Eq.~(\ref{eq:gn}) is solved directly for the step $\Delta\boldsymbol{\theta}$ at a cost that is small compared with the ODE solves. We run 15 iterations, which take about 2~min for the 150 training scenarios on a single NVIDIA GH200 GPU.

We use a second-order method because the directions in parameter space that control the biogenic correction are weakly constrained and have small gradients. In a test with station observations and noise, 400 steps of the first-order Adam optimiser from the same offline network barely changed the biogenic correction, whereas Gauss--Newton rescales these directions by the inverse curvature. Training the network online from scratch with Adam was also possible, but it took more than 20 times the computing time of the offline training followed by fine-tuning and remained less accurate.

\subsection{Results}
\label{sec:hybrid_results}

We evaluate the learned correction against the truth on a $60 \times 60$ grid over the two physical covariates of each sector, i.e. $(T, \log_{10}P)$ for the anthropogenic sector and $(T, R)$ for the biogenic sector, at three values of the synthetic covariate (0.3, 0.5, and 0.7). The grid covers the training ranges extended by the extrapolation bands. Table~\ref{tab:hybrid_rmse} lists the root-mean-square error (RMSE) of $f_s - f_s^{\mathrm{true}}$ inside the training ranges and in the extrapolation bands. For the biogenic sector, the RMSE inside the training ranges is also split into three temperature bands, because the information that the observations carry about $f_{\mathrm{bio}}$ depends strongly on temperature.

\begin{table}
    \centering
    \caption{RMSE of the learned log emission correction with respect to the truth, for the offline two-step training (offline) and after the online Gauss--Newton fine-tuning (+ online). ``Train'' denotes the training ranges of the covariates, and ``extrap.'' the extrapolation bands outside them.}
    \label{tab:hybrid_rmse}
    \small
    \begin{tabular}{llccccccc}
        \toprule
        & & \multicolumn{2}{c}{anthropogenic} & \multicolumn{5}{c}{biogenic} \\
        \cmidrule(lr){3-4} \cmidrule(lr){5-9}
        observations & method & train & extrap. & train & 5--12~$^\circ$C & 12--20~$^\circ$C & 20--30~$^\circ$C & extrap. \\
        \midrule
        all species,  & offline  & 0.063 & 0.144 & 0.162 & 0.290 & 0.062 & 0.033 & 0.315 \\
        no noise      & + online & 0.056 & 0.144 & 0.041 & 0.063 & 0.025 & 0.028 & 0.128 \\
        \addlinespace
        all species,  & offline  & 0.063 & 0.141 & 0.145 & 0.260 & 0.052 & 0.028 & 0.289 \\
        noise         & + online & 0.043 & 0.103 & 0.028 & 0.037 & 0.022 & 0.023 & 0.100 \\
        \addlinespace
        station,      & offline  & 0.063 & 0.132 & 0.287 & 0.479 & 0.145 & 0.151 & 0.544 \\
        no noise      & + online & 0.053 & 0.125 & 0.107 & 0.167 & 0.056 & 0.075 & 0.265 \\
        \addlinespace
        station,      & offline  & 0.055 & 0.121 & 0.309 & 0.528 & 0.155 & 0.127 & 0.554 \\
        noise         & + online & 0.036 & 0.106 & 0.128 & 0.212 & 0.073 & 0.062 & 0.264 \\
        \bottomrule
    \end{tabular}
\end{table}

The online fine-tuning reduces the error in every setting and region, except for the anthropogenic extrapolation error with noise-free observations of all species, which is unchanged. The anthropogenic correction is already reproduced well by the offline training, with an RMSE of about 0.06 inside the training ranges, and the fine-tuning reduces it moderately to 0.04--0.06. The largest gains are in the biogenic sector, where the RMSE inside the training ranges decreases by a factor of 2.4--5. With all species observed, the fine-tuning reduces the error in the coldest band from 0.26--0.29 to 0.04--0.06, so the biogenic correction at low temperature is constrained by the observations even though the offline training could not recover it. Low temperatures occur mainly in the early morning and on cold days, when the biogenic emissions are small but not zero, and their effect on the concentrations is therefore weak in each single scenario. A per-scenario retrieval attributes such a weak signal mostly to the prior, whereas the shared network accumulates it over all scenarios that pass through the same temperature range.

\begin{figure}
    \centering
    \includegraphics[width=\linewidth, trim=0 0 0 16, clip]{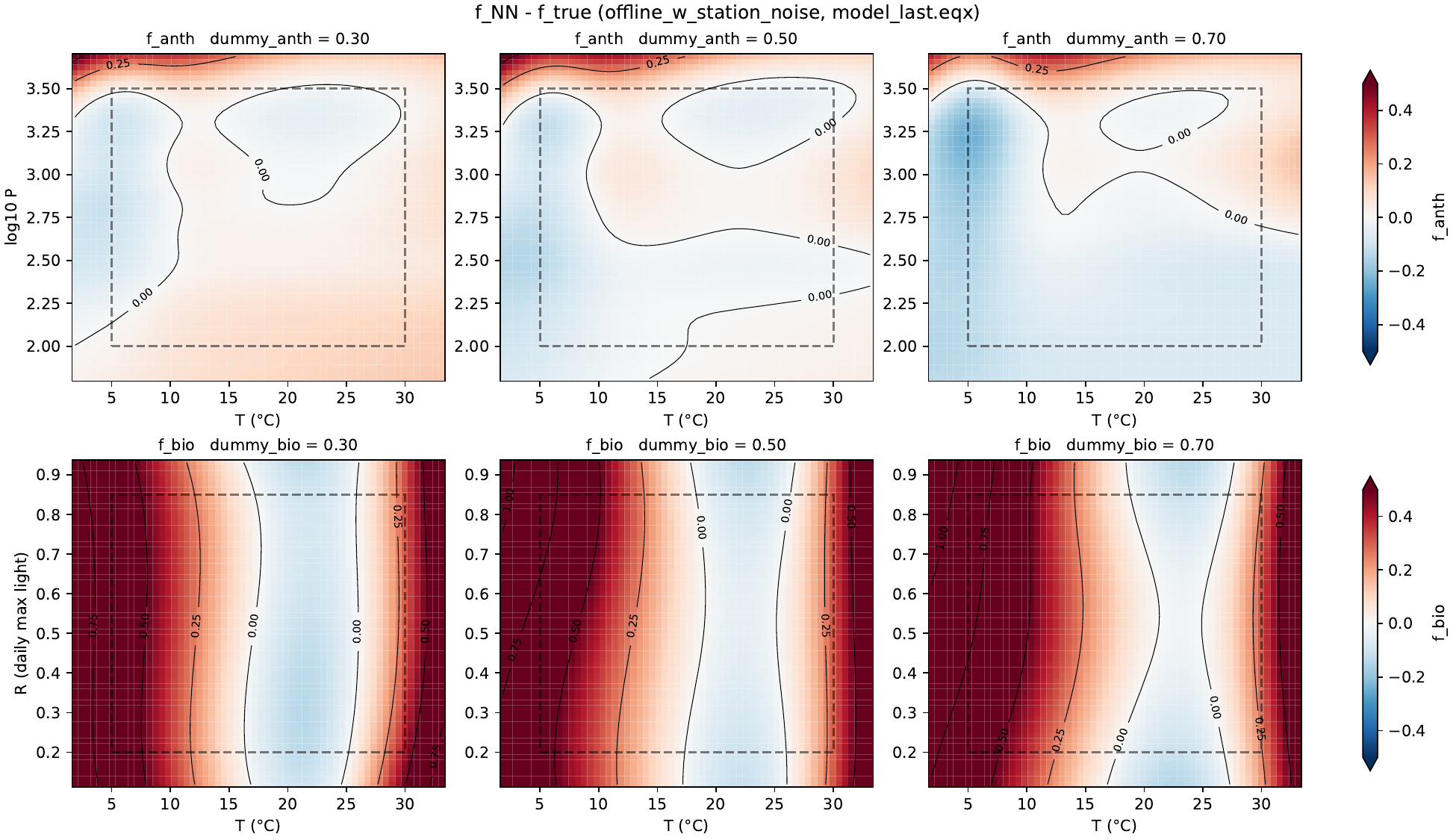}\\[0.5em]
    \includegraphics[width=\linewidth, trim=0 0 0 16, clip]{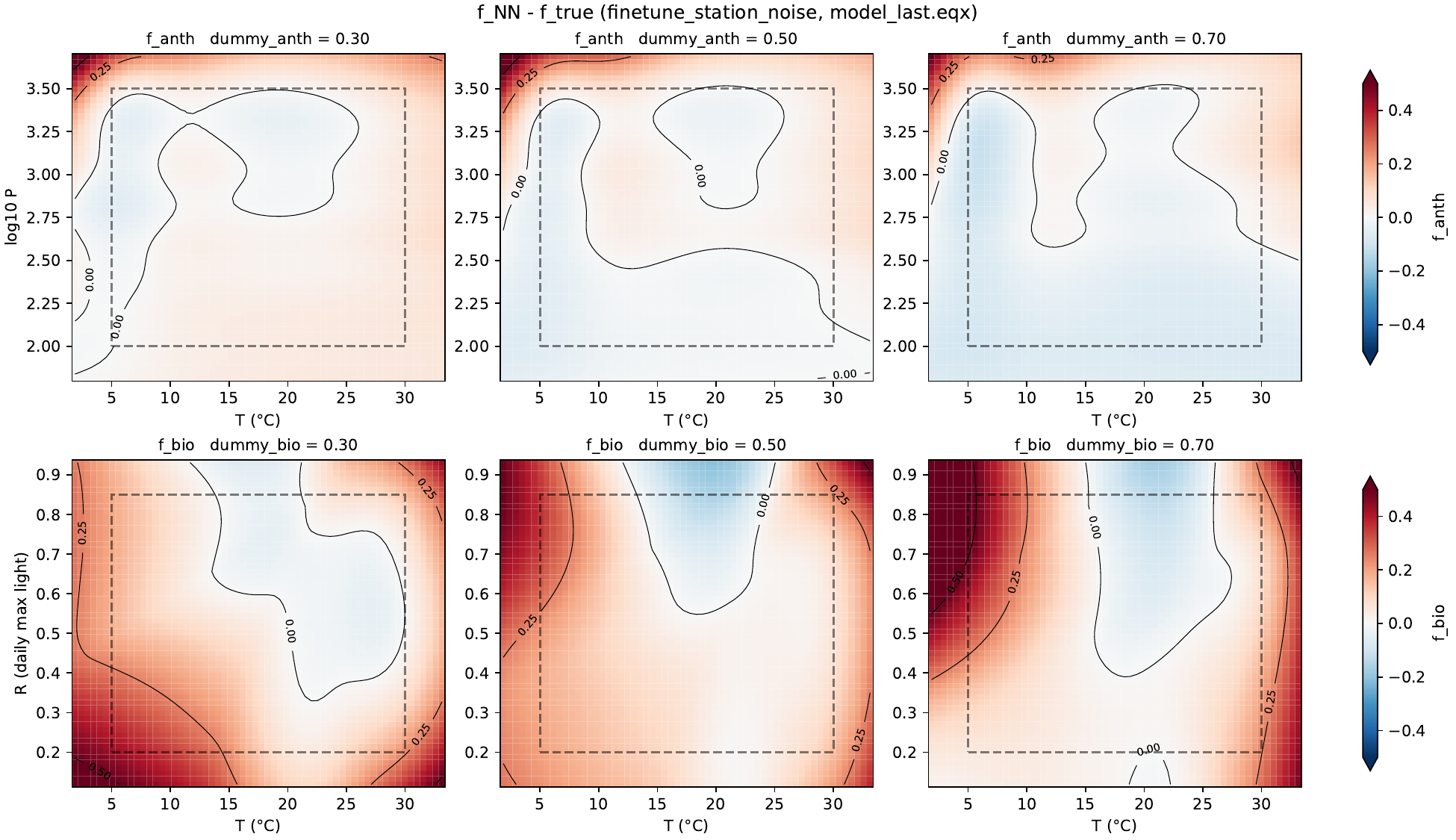}
    \caption{Error of the learned log emission correction, $f_s - f_s^{\mathrm{true}}$, with noisy station observations, (top two rows) after the offline two-step training and (bottom two rows) after the online Gauss--Newton fine-tuning. In each pair of rows, the top row shows the anthropogenic and the bottom row the biogenic sector, at three values of the synthetic covariate (columns). The dashed boxes mark the training ranges.}
    \label{fig:hybrid_error}
\end{figure}

Figure~\ref{fig:hybrid_error} shows the error of the correction for the most realistic setting, with noisy station observations. The offline network overestimates the biogenic correction both below about 12~$^\circ$C and above about 27~$^\circ$C, which means that it stays close to the inventory at both ends of the temperature range and misses both the steeper increase at low temperature and the optimum of the true response. This is at least partly a consequence of the retrievals rather than of the regression. With station observations, the evidence selects a correlation length of 192~h (Sect.~\ref{sec:hybrid_offline}), so each retrieval is nearly constant over the day and the regression cannot attribute any variation within the day to the diurnal temperature cycle. After the online fine-tuning, which evaluates the network at the instantaneous temperature of every knot, the error is mostly below 0.25 inside the training ranges, and the remaining bias is concentrated below about 10~$^\circ$C, where biogenic emissions are smallest. The anthropogenic error is small in both cases.

With all species observed, the fine-tuned biogenic error is larger without noise than with noise (0.041 and 0.028, respectively). In the noise-free setting, the iterations moved the parameters far from the offline solution along flat directions of the parameter space, to a distance $\|\boldsymbol{\theta} - \boldsymbol{\theta}_{\mathrm{off}}\| \approx 2 \times 10^6$ compared with 33--190 in the other settings, and the final iterate is not the one with the lowest validation loss. The correction remains accurate inside the training ranges, but a proximity term towards the offline parameters or early stopping would be advisable for noise-free data. Errors in the extrapolation bands are about 2--4 times larger than inside the training ranges for both approaches, which is expected for a network trained on a bounded range of covariates, and the fine-tuning reduces them in all cases but one.

\section{Discussion}
\label{sec:discussion}

The applications in Sects.~\ref{sec:param} and~\ref{sec:hybrid} use the sensitivities of the chemical model in three different ways, and the most efficient mode of automatic differentiation differs between them. When the number of parameters is small compared with the number of observations, as for the 25 rate coefficients in Sect.~\ref{sec:param} and the 50 knot values of a 4D-Var retrieval in Sect.~\ref{sec:hybrid_offline}, forward-mode differentiation through the solver gives the full Jacobian at the cost of one tangent integration per parameter and requires no checkpointing. When a scalar loss is differentiated with respect to many parameters, as in first-order training of a neural network, reverse mode with recursive checkpointing is preferable. The online Gauss--Newton fine-tuning in Sect.~\ref{sec:hybrid_online} combines both, because the network enters the ODE only through a low-dimensional set of knot values, so that the expensive sensitivities are computed in forward mode with respect to these knots, and the network parameters only enter through the Jacobian of the network itself. KPPax inherits these choices from \texttt{diffrax}, so the differentiation mode is a configurable option of the solve rather than a property of the generated code, and no adjoint has to be derived by hand for any of the three uses.

The computational cost of KPPax differs from that of KPP-generated code. Just-in-time compilation adds a fixed cost to the first call, which is amortised when the same mechanism is integrated many times, for example for many scenarios in parallel with \texttt{jax.vmap} or over many iterations of an optimisation, as in our experiments. The implicit solvers of \texttt{diffrax} solve dense linear systems with the full Jacobian of the right-hand side in each step, whereas KPP exploits the sparsity of the Jacobian through symbolic sparse LU decomposition. For POLLU, with 20 species, this difference is negligible, but the cost of dense factorisation grows with the cube of the number of species and will dominate for mechanisms with thousands of species such as the MCM. A sparse or matrix-free linear solver for the implicit stages is therefore the most important extension for large mechanisms. A systematic comparison of the accuracy and computational cost of KPPax with KPP-generated Fortran code for a large mechanism is ongoing and will be reported separately.

The log-space formulation of the rate law (Eq.~\ref{eq:rate_law_log}) keeps the rate law differentiable, but it requires strictly positive concentrations. We clip the concentrations to a small positive floor before the rate law is evaluated, which prevents the solver from probing zero or negative concentrations but also sets the gradient with respect to a clipped concentration to zero. This affects species that start from zero or are fully consumed, and applications that differentiate with respect to such near-zero concentrations should take it into account.

The identifiability analysis in Sect.~\ref{sec:param} is local, since it relies on the Jacobian at a single parameter point. It identified correctly which combinations of rate coefficients the design cannot constrain, but it underestimated how far the estimates drift along curved valleys of the cost function. Global methods, such as profile likelihoods or Markov chain Monte Carlo sampling, would quantify this drift and are straightforward to combine with KPPax, since every forward solve is differentiable and vectorisable. Because the FIM can be computed for any set of observations within seconds, it can also be used to design experiments, for example to choose the observed species, observation times, or precursor concentrations of a flow-tube experiment that make a given set of rate coefficients identifiable, before the experiment is carried out.

The emission experiment in Sect.~\ref{sec:hybrid} is idealised in several respects. The true correction is a known synthetic function, the chemical mechanism of the model is the same as that of the truth, the box model has no transport, and the initial state of each scenario is known exactly. In particular, errors in the initial state could be aliased into the correction of the early-morning emissions, where we found the largest benefit of online fine-tuning, and a joint estimation of the initial state and the emissions, as in \citet{elbern2007emission}, would be needed in a real application. The Gauss--Newton fine-tuning also forms the Hessian with respect to all network parameters, which is feasible for the 2434 parameters used here but would require matrix-free or approximate second-order methods for larger networks. Nevertheless, the experiment shows that information which is present in the observations but lost in a per-scenario retrieval can be recovered when the network is trained through the chemical model, which is only possible with a differentiable chemistry code.

\section{Conclusions}
\label{sec:conclusion}

We have presented KPPax, a kinetic preprocessor that generates differentiable chemistry code in JAX from mechanisms written in the KPP language. KPPax parses the mechanism with the \texttt{pykpp} parser, stores the stoichiometry as sparse coordinate arrays, transpiles the rate-coefficient expressions from Fortran to JAX, and evaluates the rate law with sparse operations in log space, which makes it differentiable with respect to both the concentrations and the rate coefficients. Combined with the implicit solvers and the configurable forward- and reverse-mode adjoints of \texttt{diffrax}, the generated code provides exact sensitivities of stiff chemical simulations without any hand-derived tangent-linear or adjoint code, it can be vectorised over many scenarios, and it runs on GPUs.

We demonstrated two uses of these sensitivities with the POLLU benchmark mechanism. First, the Fisher information matrix computed from a single automatically differentiated Jacobian predicted, before any fit, which rate coefficients a flow-tube-like design can constrain. Nine of the 25 rate coefficients were identifiable, and their estimates from noisy synthetic observations scattered as predicted to within about 30\%, whereas the non-identifiable rate coefficients formed clusters of fast reactions whose ratios, but not absolute values, were constrained. Second, a neural-network correction of an emission inventory, trained offline on per-scenario 4D-Var retrievals and then fine-tuned online through the chemical model by Gauss--Newton iterations, reduced the error of the biogenic correction by a factor of 2.4--5 compared with the offline training alone, at a cost of about two minutes on a single GPU. Online training recovered information on low-temperature emissions that is present in the observations but lost in the individual retrievals.

KPPax thereby brings chemical mechanisms written in the KPP language into the differentiable modelling ecosystem, in which process-based chemistry and machine learning can be combined and trained against observations. Future work will address large mechanisms, which require sparse linear solvers for the implicit integration, and applications to real chamber and flow-tube observations.

\section*{Acknowledgements}
The authors wish to acknowledge CSC -- IT Center for Science, Finland, for computational resources. WP acknowledges financial support from the Finnish Cultural Foundation, Päijät-Häme Regional Fund (grant 70251819). Large language models were used to assist with code development and with language editing of the manuscript. All code and text were reviewed by the authors, who take full responsibility for the content.

\bibliographystyle{plainnat}
\bibliography{main}

\end{document}